\documentclass[final]{nesy2026}

\usepackage{longtable}
\usepackage{booktabs}
\usepackage{placeins}
\usepackage{siunitx}
\usepackage{mathtools}
\usepackage{bm}
\usepackage[capitalise,noabbrev]{cleveref}
\usepackage{tikz}
\usetikzlibrary{arrows.meta,positioning,calc,fit,backgrounds,shapes.geometric,decorations.pathreplacing}

\RestyleAlgo{ruled}
\SetAlCapSkip{2pt}

\newcommand{\pendingfigure}[3][0.48\textwidth]{%
  \IfFileExists{#2}{\includegraphics[width=#1]{#2}}{%
    \fbox{\parbox[c][0.32\textwidth][c]{#1}{\centering\itshape%
      Figure pending\\\texttt{\detokenize{#2}}\\%
      \smallskip\scriptsize #3}}%
  }%
}

\theorembodyfont{\upshape}
\theoremheaderfont{\scshape}
\theorempostheader{:}
\theoremsep{\newline}
\newtheorem{assumption}{Assumption}

\crefname{proposition}{Proposition}{Propositions}

\newcommand{\SIGSENTENCE}{the gap is small but significant over ten
seeds (paired $t$ $p<0.05$; bootstrap $95\%$ CI $[0.009,0.053]$),
driven by Ours' low across-seed variance}

\newcommand{\Concepts}{\mathbf{c}}
\newcommand{\Inp}{\mathbf{x}}
\newcommand{\Out}{y}
\newcommand{\Onto}{\mathcal{O}}
\newcommand{\Prog}{\varphi}
\newcommand{\Parm}{\theta}
\newcommand{\Cir}{\mathcal{C}}
\newcommand{\WMC}{\mathrm{WMC}}

\newcommand{\HardAxioms}{\mathcal{H}}
\newcommand{\SoftAxioms}{\mathcal{S}}

\title[Soft Ontology Priors over Latent Concept Posteriors]%
{Soft Ontology Priors over Latent Concept Posteriors\\
for Weakly Supervised Neuro-Symbolic Learning}

\clearauthor{\Name{Tengwei Song}\and
  \Name{Robert Hoehndorf}\\
  \addr CEMSE, King Abdullah University of Science and Technology,
  Thuwal, Saudi Arabia\\
  \normalfont\Email{\{tengwei.song,robert.hoehndorf\}@kaust.edu.sa}}

\begin{document}

\maketitle

\begin{abstract}
Neuro-symbolic (NeSy) predictors that supervise only the task label
suffer from \emph{reasoning shortcuts}: distinct concept mappings
achieve the same maximum likelihood, leaving the recovered concepts
unidentifiable and unsafe for downstream reuse. We attack this failure
by injecting a \emph{soft ontology prior} over the latent concepts
into the training objective and compiling it together with the task
constraint into a smooth, decomposable probabilistic circuit. Under
independent Bernoulli perception, the normalisers of the resulting
structured concept posterior reduce to weighted model counts that are
linear in the compiled circuit size, because the neural leaf weights
fold into the circuit before evaluation. The soft compilation strategy places
\emph{expert} axioms in the circuit's \emph{weights} rather than its
\emph{support} (task-structural axioms guaranteed by the
data-generating process stay hard), so noisy expert axioms only
redistribute prior mass and cannot empty the support. We characterise how the prior reweights label-restricted shortcut
mass through differences in ontology violation energy, show that soft priors incur
bounded log-density perturbations under bounded axiom noise, and prove
that the hard-support counterpart can be made degenerate by a single
corrupted axiom. We validate these predictions on an audited
six-concept view of the real-world BDD-OIA driving benchmark and on
controlled NeSy tasks.
\end{abstract}


\section{Introduction}
\label{sec:intro}

Neuro-symbolic (NeSy) AI integrates neural perception with symbolic
reasoning to build interpretable, constraint-aware
predictors~\citep{manhaeve2018deepproblog,xu2018semanticloss,serafini2016ltn}.
Much of their appeal rests on the learned concepts being trustworthy:
once the model recovers human-readable concepts, those concepts can be
reused, verified and explained. Recent work, however, shows that this
promise fails systematically. NeSy predictors trained on task labels
alone can reach near-perfect accuracy by acquiring \emph{reasoning
shortcuts}~\citep{marconato2023notall,vankrieken2024independence}:
concept maps that satisfy the symbolic program for the wrong reasons.
Reasoning shortcuts are invisible at the label level, break
out-of-distribution generalisation, and undermine the safety story
that motivates NeSy in the first place.

Two recent lines of work address reasoning shortcuts from the
perception side. BEARS stabilises predictions by ensembling concept
extractors with a repulsive KL
term~\citep{marconato2024bears}. NeSy-DM replaces the one-shot
factorised posterior with a masked diffusion model that can express
concept correlations~\citep{vankrieken2025nesydm}. Both interventions
act on the perception model while leaving the training objective
unchanged, and neither offers a principled way to inject existing
domain knowledge into the loss surface itself.

We take the complementary route: we leave the perception model
factorised and instead modify the training objective so that it is
informed by a soft, axiom-level prior over the latent concepts. A
parallel line of work, the ontology-to-circuit pipeline of
\citet{lazzari2026ontopc}, shows that a description-logic ontology can
be compiled into a probabilistic circuit and used to guarantee that
\emph{supervised} multi-label classifiers respect every axiom; that
work targets ontology-consistent output prediction, not the
latent-concept NeSy regime in which only task labels are observed.
We use this compilation idea in a different regime: the ontology
defines a soft prior over hidden concepts inside the task-label
likelihood, rather than a hard support constraint on observed
classifier outputs. The prior is
soft (axioms downweight rather than delete worlds), so the resulting
objective degrades gracefully when the ontology is noisy or incomplete,
in contrast to a hard support restriction that can render the joint
program unsatisfiable on a single mislabelled axiom.

\paragraph{Contributions.}
We make three contributions. First, we introduce a soft ontology prior over latent concepts
for NeSy predictors trained from task labels only, and incorporate it into the structured
concept posterior used in the training objective. Second, we derive a normalised WMC objective that combines task constraints,
neural evidence, and ontology-derived axiom weights in a compiled circuit with linear-time
evaluation, and show how the prior reweights label-preserving shortcut
maps before deterministic collapse. Third, we prove that soft and hard ontology priors behave differently under axiom
corruption: bounded perturbations induce bounded log-density changes for soft priors,
while hard-support constraints can collapse after one corrupted axiom. Experiments on
real-world and controlled NeSy benchmarks support these predictions and show that gains
require ontology information that resolves ambiguity left by task-label supervision.

\section{Related Work}
\label{sec:related}
\paragraph{Reasoning shortcuts in neuro-symbolic learning.}
The notion of reasoning shortcut was formalised by
\citet{marconato2023notall}, who prove a counting bound on the number
of label-equivalent concept maps and identify four root causes;
\citet{vankrieken2024independence} further show that the factorised
concept posterior shared by DeepProbLog~\citep{manhaeve2018deepproblog},
the semantic loss~\citep{xu2018semanticloss},
A-NeSI~\citep{vankrieken2023anesi} and
SPL~\citep{ahmed2022spl} is structurally biased toward deterministic
concept maps. BEARS~\citep{marconato2024bears} mitigates this bias
with a KL-repulsion ensemble of factorised concept extractors, while
NeSy-DM~\citep{vankrieken2025nesydm} replaces the one-shot
factorised posterior with a masked discrete diffusion model that can
express concept correlations. Both interventions act on the
perception architecture; we instead modify the training objective by
folding a soft ontology prior into the weighted circuit, leaving the
factorised perception model untouched.

\paragraph{Probabilistic circuits for symbolic constraints.}
Smooth, decomposable circuits admit linear-time weighted model
counting~\citep{darwiche2001dnnf,darwiche2002kc,chavira2008wmc}; under
compatible scopes or vtrees, the same circuit family is closed under
products and conditioning~\citep{choi2020pc,vergari2021atlas}. This
makes them a standard backbone for tractable neuro-symbolic
inference. SPL~\citep{ahmed2022spl} compiles the task constraint into
such a circuit and trains the perception model against the resulting
exact likelihood; \citet{lazzari2026ontopc} extend the same machinery
to entire description-logic ontologies, but use the compiled circuit
to guarantee ontology consistency at the output of a \emph{supervised}
multi-label classifier. Our work targets the complementary setting in
which the concepts themselves are latent and only the task label is
observed: the compiled ontology serves as a soft prior over hidden
concepts inside the training objective, rather than as a hard
projection on the output.

\paragraph{Posterior regularisation and constrained inference.}
Our soft prior also relates to posterior
regularisation~\citep{ganchev2010posterior} and generalised expectation
criteria~\citep{mann2010generalized}, which inject side-information into
latent-variable learning by penalising posterior moments against
user-specified targets, treating the constraint as a soft auxiliary
term. We instead compile the constraint set into a weighted circuit and
absorb the structured posterior into a single exact normalised
likelihood, enforcing the constraint at the partition function rather
than as a moment regulariser; this avoids moment-matching and recovers
exact gradients through the ontology circuit by leaf reweighting.

\paragraph{Constraint-driven neural learning.}
MultiplexNet~\citep{hoernle2022multiplexnet}, projection
fixers~\citep{stoian2023ccn}, semantic-based
regularisation~\citep{diligenti2017semreg}, Logic Tensor
Networks~\citep{badreddine2022ltn} and PyLon~\citep{ahmed2023pylon}
treat symbolic knowledge as an output-level constraint or
regulariser rather than a prior over latent concepts. Concept bottleneck
models~\citep{koh2020cbm} motivate treating concepts as first-class; the
restricted axiom language used in this paper, derived from EL-style
inclusions with small Boolean and cardinality macros, delimits our
compiler's tractable target~\citep{baader2005el,baader2017dlhandbook}.

\section{Preliminaries and Problem Setting}
\label{sec:prelim}

\paragraph{Latent-concept NeSy predictors.}
We consider tasks in which an input $\Inp\in\mathcal{X}$ is mapped to
a \emph{latent concept vector} $\Concepts\in\{0,1\}^n$ by a neural
perception model $p_{\Parm}(\Concepts\mid\Inp)$ and a fixed symbolic
program $\Prog:\{0,1\}^n\!\to\!\mathcal{Y}$ produces the observable
output $\Out=\Prog(\Concepts)$. Following standard NeSy
practice~\citep{manhaeve2018deepproblog,xu2018semanticloss},
$p_{\Parm}$ is factorised as $p_{\Parm}(\Concepts\mid\Inp)=
\prod_{i=1}^{n} p_{\Parm}(c_i\mid\Inp)$ and trained by maximising the
\emph{semantic loss} $\log p_{\Parm}(\Out\mid\Inp)=
\log\sum_{\Concepts:\Prog(\Concepts)=\Out}p_{\Parm}(\Concepts\mid\Inp)$.
A \emph{reasoning shortcut}~\citep[Def.~1]{marconato2023notall} is a
non-identity deterministic map $\alpha:\Concepts\mapsto\Concepts'$ with
$\Prog(\alpha(\Concepts))=\Prog(\Concepts)$ on the support of the
data-generating concept distribution $p^\star$; we write $\mathcal{A}$
for the set of all such label-preserving maps.
\citet{vankrieken2024independence} further show that the
factorised optimum concentrates on deterministic Bernoullis, so the
loss surface is a disconnected union of vertex flats and gradient
descent selects one by the random initialisation.

\paragraph{Probabilistic circuits and the EL ontology fragment.}
A \emph{probabilistic circuit}~\citep{choi2020pc} is a rooted DAG with
\textsc{Leaf}, \textsc{Product} and \textsc{Sum} nodes; under
decomposability, smoothness and determinism (sd-DNNF) it admits a
linear-time weighted model count
$\WMC(\Cir;w)=\sum_{\Concepts}w(\Concepts)\mathbf{1}[\Concepts\models\Cir]$
by a single bottom-up
pass~\citep{darwiche2001dnnf,darwiche2002kc,chavira2008wmc}. SPL
\citep{ahmed2022spl} exploits this in NeSy by compiling the task
constraint $\Prog(\Concepts)=\Out$ into a circuit
$\Cir_{\Prog}^{\Out}$ whose WMC under independent Bernoulli leaf
weights equals the semantic loss. We compile a restricted ontology language built from EL-style
inclusions together with small Boolean and cardinality macros,
including \textsc{SubClass}, \textsc{Implies}, \textsc{Disjoint}, and
\textsc{ExactlyOne}; this covers task-induced one-hot schemas,
benchmark-provided binary CNF priors, and expert traffic rules over
audited concept views \citep{baader2005el,baader2017dlhandbook}.

\paragraph{Worked two-concept example.}
Let $\Concepts=(r,g)$ be a red and green light with evidence
$\theta_r,\theta_g\in[0,1]$ and axiom $\textsc{Disjoint}(r,g)$ of weight
$\lambda$. The hard variant deletes the world $(1,1)$; the soft variant
keeps it with factor $e^{-\beta\lambda}$, leaving the other three worlds
unchanged. Write $Z_y$ for the label-restricted partition later defined in
\Cref{sec:method-posterior}. For $\Out=\texttt{stop}$ ($\Prog(r,g)=r$),
$Z_y^{\textsc{hard}}=\theta_r(1{-}\theta_g)$ and
$Z_y^{\textsc{soft}}=Z_y^{\textsc{hard}}+\theta_r\theta_g\,e^{-\beta\lambda}$.
If a further mislabelled axiom also rules out $(0,0)$, the hard $Z_y$ of
any label requiring $(0,0)$ collapses to $0$ while the soft variant only
rescales the violating leaf, the failure mode that
Propositions~\ref{prop:count} and~\ref{prop:noise} formalise.

\section{Circuit-Regularized Concept Posteriors}
\label{sec:method}

\begin{figure}[t]
\centering
\includegraphics[width=\textwidth]{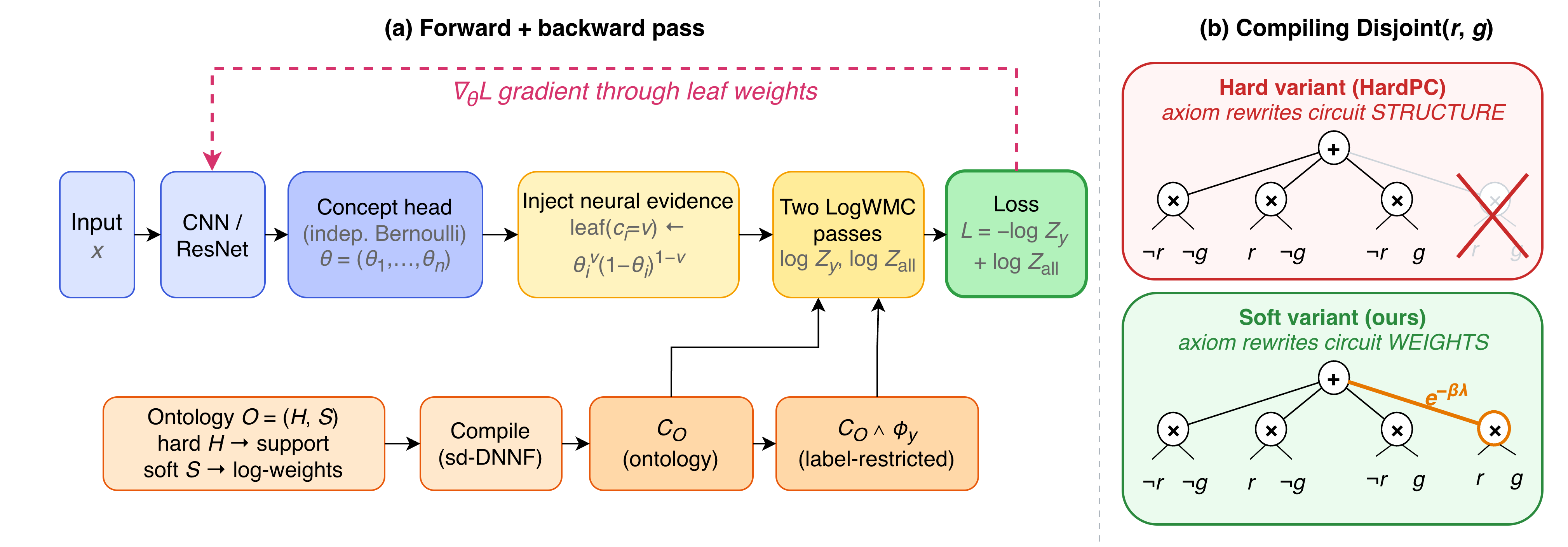}
\caption{Method overview. (a) Forward + backward pass with leaf
reweighting and two LogWMC evaluations. (b) Compiling one
$\mathsf{Disjoint}(r,g)$ axiom: hard support removal (left) vs.\
soft edge reweighting (right).}
\label{fig:overview}
\end{figure}

\subsection{Soft Ontology Prior}
\label{sec:method-prior}
Let $\Onto=(\HardAxioms,\SoftAxioms)$ be a finite set of axioms
drawn from the restricted language of \S\ref{sec:prelim}, defined
over the concept vocabulary $\{c_1,\dots,c_n\}$, together with a
provenance tag indicating whether the axioms come from a benchmark
CNF, a task schema, or an expert-authored mapping. To keep the hard/soft
split stable across methods we tag axioms by provenance:
$\HardAxioms_{\mathrm{task}}$ (task-structural axioms guaranteed by the
data-generating process, e.g.\ the MNIST-Add one-hot positional schema,
always hard);
$\SoftAxioms_{\mathrm{bench}}$ (benchmark-provided distribution priors
that are not satisfied by every training sample, e.g.\ the
MNLogic in-distribution CNF whose benchmark configuration draws
out-of-distribution combinations with positive probability, soft for
our method);
$\SoftAxioms_{\mathrm{expert}}$ (expert axioms over audited views,
e.g.\ the BDD-OIA-6c traffic rules, soft for our method);
and $\HardAxioms_{\mathrm{baseline}}=\HardAxioms_{\mathrm{task}}\cup
\SoftAxioms_{\mathrm{bench}}\cup\SoftAxioms_{\mathrm{expert}}$, used by
HardPC$^\star$, which hardens every benchmark and expert axiom into the
support. Soft axioms only
down-weight worlds; each soft $a_j\in\SoftAxioms$ contributes a
non-negative cost
$\lambda_j\,v_j(\Concepts)$, where $v_j(\Concepts)\in[0,1]$ is zero
exactly when $\Concepts$ satisfies $a_j$; we write
$E_{\Onto}(\Concepts)=\sum_{j}\lambda_j v_j(\Concepts)$ for the resulting
per-world \emph{ontology energy}.  We use the unnormalised ontology score
\begin{equation}
  \widetilde p_{\Onto}(\Concepts;\beta) \;=\;
  \mathbf{1}[\Concepts\models\HardAxioms]\,
  \exp\!\Bigl(-\beta\sum_{j=1}^{|\SoftAxioms|}\lambda_j v_j(\Concepts)\Bigr),
  \label{eq:soft-prior}
\end{equation}
with $\beta\!\ge\!0$ controlling prior strength.  $\beta\!=\!0$ ignores
soft penalties while preserving hard support; $\beta\!\to\!\infty$
turns every positive soft violation cost into a hard exclusion, matching
the hard-constraint regime studied by
\citet{hoernle2022multiplexnet,stoian2023ccn}.  After multiplication by
$p_{\Parm}(\Concepts\mid\Inp)$, the normaliser of the product
$p_{\Parm}\widetilde p_{\Onto}$ depends on $\Parm$, so the likelihood
below keeps both partitions: the label-restricted $Z_y$ and all-label
$Z_{\mathrm{all}}$.  Ontology costs live in a weighted circuit:
leaf weights take neural evidence; internal structure and fixed
log-weights encode hard support and soft penalties.

\paragraph{Structured concept posterior.}
\label{sec:method-posterior}
Combining the task constraint, the perception model and the prior
yields the \emph{structured concept posterior}
$p(\Concepts\mid\Inp,\Out,\Onto)\propto
\mathbf{1}[\Prog(\Concepts)=\Out]\,p_{\Parm}(\Concepts\mid\Inp)\,
\widetilde p_{\Onto}(\Concepts;\beta)$, with label-restricted
$Z_y(\Inp)=\sum_{\Concepts:\Prog(\Concepts)=\Out}
p_{\Parm}(\Concepts\mid\Inp)\widetilde p_{\Onto}(\Concepts;\beta)$ and
all-label $Z_{\mathrm{all}}(\Inp)=
\sum_{\Concepts}p_{\Parm}(\Concepts\mid\Inp)
\widetilde p_{\Onto}(\Concepts;\beta)$. They define the conditional
probability $p(\Out\mid\Inp,\Onto)=Z_y(\Inp)/Z_{\mathrm{all}}(\Inp)$
whose normalised log-likelihood
\begin{equation}
  \mathcal{L}(\Parm) \;=\; \sum_{(\Inp,\Out)\in\mathcal{D}}
  \bigl[\log Z_y(\Inp)-\log Z_{\mathrm{all}}(\Inp)\bigr]
  \label{eq:loss}
\end{equation}
is our training objective; the unnormalised surrogate
$\sum\log Z_y$ is improper unless $Z_{\mathrm{all}}$ is constant in
$\Parm$ and we do not use it.

\subsection{Tractable Compilation, Training, and Identifiability}
\label{sec:method-compile}
Direct enumeration of $Z_y$ and $Z_{\mathrm{all}}$ is exponential in
$n$. The combination of an independent Bernoulli $p_{\Parm}$ and
compiled weighted circuits makes both WMCs linear in the circuit sizes
actually evaluated.

\begin{assumption}[Compatible compilation]
\label{ass:compat}
For each label $\Out$, the circuit $\Cir^{\Out}_{\Onto,\Prog}$ is either
(i) a jointly compiled sd-DNNF for
$\HardAxioms\land[\Prog(\Concepts)=\Out]$ carrying the soft weights of
$\widetilde p_{\Onto}$, or (ii) a product of $\Cir_{\Onto}$ and the SPL
task circuit over a shared vtree that is itself smooth, decomposable and
deterministic. We do \emph{not} assume arbitrary circuit products are
tractable; Appendix~\ref{apd:compile} gives a constructive compilation that
meets this for the restricted axiom language and task programs used here, with its
size.
\end{assumption}

\begin{proposition}[Evidence folding]
\label{thm:leaf-fold}
Let $\Cir_{\Onto}$ be a smooth, decomposable weighted circuit whose
WMC under unit neural evidence equals
$\sum_{\Concepts}\widetilde p_{\Onto}(\Concepts;\beta)$, and let
$\Cir^{\Out}_{\Onto,\Prog}$ satisfy \Cref{ass:compat}. Inject neural
evidence by multiplying every literal $(c_i\!=\!v)$ by
$\theta_i^{v}(1-\theta_i)^{1-v}$. Then the root WMCs equal $Z_y(\Inp)$
and $Z_{\mathrm{all}}(\Inp)$, and the loss in \Cref{eq:loss} is
computable in
$\mathcal{O}\bigl(|\Cir^{\Out}_{\Onto,\Prog}|+\allowbreak
|\Cir_{\Onto}|\bigr)$, linear in
the size of the materialised circuits, not of the unconjoined ontology
circuit (proof in Appendix~\ref{apd:proofs}).
\end{proposition}

With the circuits compiled once per $(\Onto,\Prog)$ pair, each
training step costs only neural-evidence injection plus two
bottom-up $\mathrm{logsumexp}$ WMCs by
Proposition~\ref{thm:leaf-fold}; the corresponding pseudocode is
Algorithm~\ref{alg:forward} in Appendix~\ref{apd:compile}.

\paragraph{Prior weighting of shortcut maps.}
Each shortcut $\alpha\in\mathcal{A}$ acquires an \emph{ontology violation
energy} $V_{\Onto}(\alpha)=\mathbb{E}_{\Concepts\sim p^\star}
[\sum_{j}\lambda_j v_j(\alpha(\Concepts))]$ under the soft prior, i.e.\
its expected weighted soft-axiom violation on the true distribution.
Proposition~\ref{prop:count} isolates how the ontology reweights
label-restricted mass, equivalently the structured posterior before
exact deterministic collapse.

\begin{proposition}[Shortcut reweighting]
\label{prop:count}
Assume the setting of prior shortcut
analysis~\citep[Thm.~2]{marconato2023notall}. In particular,
$p_{\Parm}$ collapses to a single assignment $\alpha(\Concepts)$ per
input, and empirical sums are replaced by expectations under
$p^\star$. Then under
the soft score $\widetilde p_{\Onto}(\cdot;\beta)$ of
\Cref{eq:soft-prior}, for any two label-preserving shortcut maps
$\alpha_1,\alpha_2\in\mathcal{A}$ the ratio of their label-restricted
ontology-weighted scores
$S_\Onto(\alpha)=
\exp[-\beta\,V_{\Onto}(\alpha)]$ is
\[
  \frac{S_\Onto(\alpha_1)}{S_\Onto(\alpha_2)}
  = \exp\!\bigl(-\beta[V_{\Onto}(\alpha_1)-V_{\Onto}(\alpha_2)]\bigr).
\]
The normalised objective in \Cref{eq:loss} uses this factor away from
the exact deterministic limit, but in that limit
$Z_y=Z_{\mathrm{all}}$ for every label-preserving point-mass map. In
the hard limit $\beta\!\to\!\infty$, maps with positive violation
energy have zero label-restricted support.
\end{proposition}

\begin{proposition}[Bounded log-density perturbation; hard collapse]
\label{prop:noise}
Let $\Onto'$ be obtained from $\Onto$ by an axiom-corruption model
that changes at most a fraction $\rho$ of the $K$ soft axiom weights,
with each changed axiom moving any world's energy by at most $\Delta$.
Equivalently, assume the bounded per-world energy perturbation
$|E_{\Onto}(\Concepts)-E_{\Onto'}(\Concepts)|\le\varepsilon$ with
$\varepsilon=\mathcal{O}(\rho K\Delta)$ ($K$ axioms, $\Delta$ maximum
per-axiom gap, $\rho$ flipped fraction). Then
$|\log p_{\Onto}(\Concepts;\beta)-\log p_{\Onto'}(\Concepts;\beta)|
\le 2\beta\varepsilon$ for every world, so
$\mathrm{KL}\!\bigl(p_{\Onto}(\cdot;\beta)\|p_{\Onto'}(\cdot;\beta)\bigr)
\le 2\beta\varepsilon$. The hard variant
$\HardAxioms_{\mathrm{baseline}}$ instead suffers
$\mathrm{KL}=\infty$ and a vanishing gradient on $p_{\Parm}$ as soon
as a flipped axiom excludes a world with $p^\star(\Concepts)>0$.
\end{proposition}

\section{Experiments}
\label{sec:exp}

We evaluate whether soft ontology priors improve latent-concept
recovery, when the improvement appears, and how the method behaves
under corrupted ontologies. BDD-OIA-6c tests concept recovery on an
audited real-world view, while MNIST-Add isolates the role of an
informative prior in a controlled digit-sum task. MNLogic-3bit-CNF
tests whether benchmark-provided CNF priors can be used as soft
distribution priors, and MNIST-Half checks the same digit-sum setting
under sparse training. Ablations vary prior richness, prior strength,
normalisation, and hard versus soft axiom use.

\subsection{Experimental setup}
\label{sec:exp-setup}

\paragraph{Benchmarks.}
\textbf{BDD-OIA-6c}~\citep{xu2020bddoia}: the audited canonical
traffic view derived from the RSBench preprocessed
release~\citep{bortolotti2024rsbench}; $\Concepts\in\{0,1\}^6$ and a
binary stop/forward label whose deterministic rule $\Prog(\Concepts)$
and concept vocabulary are listed in Appendix~\ref{apd:data}.
\textbf{MNIST-Add}~\citep{manhaeve2018deepproblog}: pairs of MNIST
digit images observed only by their sum;
$\Concepts\in\{0,1\}^{20}$ one-hot over two ten-digit positions,
$\Prog(\Concepts)=d_1+d_2\in\{0,\dots,18\}$.
\textbf{MNIST-Half}~\citep{marconato2024bears}: sparse-training
adaptation of MNIST-Add over a five-digit alphabet;
$\Concepts\in\{0,1\}^{10}$, $\Prog(\Concepts)=d_1+d_2$. Training
pairs are restricted to the unordered set
$\{(0,0),(0,1),(2,3),(2,4)\}$ (four distinct sums); the remaining
$18/25$ ordered pairs in $\{0,\dots,4\}^2$ form an OOD split.
\textbf{RSBench MNLogic-3bit-CNF}~\citep{bortolotti2024rsbench}:
three binary digits $\Concepts=(a,b,d)\in\{0,1\}^3$,
$\Prog(\Concepts)=(a\wedge b)\vee\neg d$, with the official
in-distribution combinations $\{001,010,000\}$ encoded as a DIMACS
CNF prior. Per-dataset split sizes, ontology level definitions for
MNIST-Add and MNIST-Half, the BDD-OIA-6c stop rule, and the MNLogic
DIMACS encoding are all in Appendix~\ref{apd:data}.
\paragraph{Baselines.}
\emph{DPL$^\ddagger$}~\citep{manhaeve2018deepproblog}: $\ddagger$~replaces the original categorical head with a factorised Bernoulli for comparability.
\emph{SPL$^\dagger$}~\citep{ahmed2022spl}: $\dagger$~adds an explicit concept head so that latent concepts are evaluable in our weakly supervised setting.
\emph{HardPC$^\star$}~\citep{lazzari2026ontopc}: $\star$~uses the same compiled ontology rules as hard support constraints, adapted from the original supervised setting to our weakly supervised setting.
\emph{BEARS$^\S$}~\citep{marconato2024bears}: published $k\!=\!3$ ensemble has $3\times$ the parameters of all other methods; $\S$~uses the structured concept argmax of a single member at inference.
We additionally evaluate BEARS$^\S$(k\!=\!1) for a parameter-matched comparison.
\emph{Oracle} provides the upper bound with full concept supervision.

\paragraph{Metrics.}
Our primary concept metric is macro F1 (F1(C)) on the raw-$\theta$
perception output: this evaluates the learned perception head
directly, rather than crediting concept assignments recovered only
by label-conditioned symbolic decoding, and we use it consistently
in all main tables. We also report task accuracy and concept
ECE~(\citealp{naeini2015ece,guo2017calibration}) on the raw-$\theta$
head. The multi-seed identifiability diagnostic in
\Cref{sec:exp-main} additionally reports F1 on the
\emph{structured} concept decoder—the circuit MAP argmax given the
model's predicted task label—to separate perception-head quality
from the symbolic layer's decoding capability.

\paragraph{Protocol.}
BDD-OIA-6c uses a two-layer MLP head over frozen $2048$-d features;
MNIST-Add and MNIST-Half use a two-position CNN concept extractor;
MNLogic-3bit-CNF uses a three-position CNN concept extractor. All
benchmarks train with Adam ($\mathrm{lr}=10^{-3}$) and use
$\beta\!=\!1$ for Ours. MNIST-Half trains for $60$ epochs following
the BEARS sparse-training protocol; BDD-OIA-6c, MNIST-Add, and
MNLogic train for $30$ epochs. Batch size is $256$ for BDD-OIA-6c
and $128$ for the three MNIST-style benchmarks.
Tables and figures report mean $\pm$ std over ten random seeds.
Full dataset specifications,
audits and hyper-parameters are
in Appendices~\ref{apd:data} and~\ref{apd:impl}; all runs use a single NVIDIA A100
($40$\,GB) on a 32-core AMD EPYC 7513 node. Appendix~\ref{apd:compile}
reports the materialised circuit sizes. Code and experiment
configurations are available at
\url{https://github.com/tewiSong/circuit-nesy}.

\subsection{Main benchmark results}
\label{sec:exp-main}

\Cref{tab:main} and \Cref{tab:mnist-add} give the main comparisons.
On BDD-OIA-6c, all non-Oracle methods reach similar task accuracy,
but Ours gives the best non-Oracle raw concept F1. The gain is small
but consistent: \SIGSENTENCE. The low variance was computed over the
same ten seeds as all baselines and reflects that the prior steers
runs toward the same audited traffic view; the remaining Oracle gap
shows that shortcut behaviour is only mitigated. MNLogic is less
discriminative because its three-bit task leaves several concept maps
nearly exchangeable, yielding high seed variance, but it is consistent
with treating benchmark CNF priors as soft distribution priors when
they are not guaranteed constraints.

\begin{table}[t]
\centering
\caption{\textbf{Main benchmark results.}
Left: BDD-OIA-6c, the audited real-world concept view.
Right: MNLogic-3bit-CNF, used to test benchmark-provided CNF priors.
F1(C) and ECE are computed on the raw-$\theta$ concept head.
\textbf{Bold}/\underline{underline} $=$ best/second-best non-Oracle
per column.}
\label{tab:main}
\resizebox{\linewidth}{!}{%
\begin{tabular}{l ccc ccc}
\toprule
& \multicolumn{3}{c}{BDD-OIA-6c} & \multicolumn{3}{c}{MNLogic-3bit-CNF} \\
\cmidrule(lr){2-4}\cmidrule(lr){5-7}
Method & F1(C)\,$\uparrow$ & Acc.(Y)\,$\uparrow$ & ECE\,$\downarrow$
       & Acc.\,$\uparrow$  & F1(C)\,$\uparrow$ & ECE\,$\downarrow$\\
\midrule
DPL$^\ddagger$    & $0.309 \pm 0.033$                   & $0.766 \pm 0.004$                   & $0.346 \pm 0.024$                   & $0.9989\pm0.0008$                   & $0.775\pm0.202$                     & $\mathbf{0.105\pm0.091}$\\
SPL$^\dagger$     & $0.279 \pm 0.049$                   & $0.766 \pm 0.002$                   & $0.331 \pm 0.019$                   & $\underline{0.9999\pm0.0002}$       & $\underline{0.776\pm0.231}$         & $0.111\pm0.123$\\
HardPC$^\star$    & $0.330 \pm 0.040$                   & $0.766 \pm 0.004$                   & $0.340 \pm 0.011$                   & $0.9997\pm0.0001$                   & $0.520\pm0.004$                     & $0.441\pm0.009$\\
BEARS$^\S$ (k\!=\!3) & $0.291 \pm 0.017$              & $\underline{0.767 \pm 0.002}$       & $\mathbf{0.184 \pm 0.001}$          & $0.9997\pm0.0004$                   & $0.521\pm0.058$                     & $0.177\pm0.061$\\
BEARS$^\S$ (k\!=\!1) & $\underline{0.330 \pm 0.036}$  & $0.765 \pm 0.003$                   & $\underline{0.326 \pm 0.012}$       & $0.9997\pm0.0006$                   & $0.728\pm0.240$                     & $0.143\pm0.127$\\
Ours       & $\mathbf{0.362 \pm 0.001}$          & $\mathbf{0.768 \pm 0.002}$          & $0.351 \pm 0.003$                   & $\mathbf{0.9999\pm0.0001}$          & $\mathbf{0.780\pm0.234}$            & $\underline{0.109\pm0.125}$\\
\midrule
Oracle     & $0.456 \pm 0.003$          & $0.770 \pm 0.002$          & $0.126 \pm 0.001$          & $0.9999\pm0.0001$          & $1.000\pm0.001$          & $0.000\pm0.000$\\
\bottomrule
\end{tabular}%
}
\end{table}

MNIST-Add gives the clearest controlled effect (\Cref{tab:mnist-add}).
The level-$3$ prior supplies the left/right ordering information that
the sum label does not identify, and this produces the largest
concept-F1 improvement. This prior is deliberately task-specific: it
helps because the constructed data-generating process fixes an order
that the sum label hides; when that information is removed (L2/L1),
the gains shrink. MNIST-Half gives a sparse-training check in the same
digit-sum family and is reported alongside MNIST-Add for compactness.

\begin{table}[t]
\centering
\caption{\textbf{Digit-sum results.}
Left: MNIST-Add, the controlled setting for testing informative
priors. Right: MNIST-Half, following the
BEARS sparse-training variant.}
\label{tab:mnist-add}
\resizebox{\linewidth}{!}{%
\begin{tabular}{l ccc ccc}
\toprule
& \multicolumn{3}{c}{MNIST-Add (full sums)} & \multicolumn{3}{c}{MNIST-Half (BEARS sparse training)}\\
\cmidrule(lr){2-4}\cmidrule(lr){5-7}
Method & Acc.\,$\uparrow$ & F1(C)\,$\uparrow$ & ECE\,$\downarrow$
       & Acc.\,$\uparrow$ & F1(C)\,$\uparrow$ & ECE\,$\downarrow$\\
\midrule
DPL$^\ddagger$        & 0.526 $\pm$ 0.124          & 0.215 $\pm$ 0.110          & \underline{0.137 $\pm$ 0.024} & 1.000 $\pm$ 0.001          & 0.278 $\pm$ 0.054          & 0.233 $\pm$ 0.028\\
SPL$^\dagger$         & 0.985 $\pm$ 0.005          & 0.213 $\pm$ 0.076          & 0.169 $\pm$ 0.015          & 1.000 $\pm$ 0.000          & 0.272 $\pm$ 0.007          & \underline{0.232 $\pm$ 0.006}\\
HardPC$^\star$        & 0.987 $\pm$ 0.002          & \underline{0.267 $\pm$ 0.070} & 0.225 $\pm$ 0.029        & 1.000 $\pm$ 0.000          & \underline{0.298 $\pm$ 0.042} & 0.286 $\pm$ 0.024\\
BEARS$^\S$ (k\!=\!1)  & 0.985 $\pm$ 0.005          & 0.212 $\pm$ 0.131          & 0.170 $\pm$ 0.022          & 1.000 $\pm$ 0.000          & 0.271 $\pm$ 0.006          & 0.233 $\pm$ 0.005\\
BEARS$^\S$ (k\!=\!3)  & \textbf{0.991} $\pm$ 0.002 & 0.240 $\pm$ 0.093          & \textbf{0.112} $\pm$ 0.043 & \textbf{1.000} $\pm$ 0.000 & 0.296 $\pm$ 0.076          & \textbf{0.118} $\pm$ 0.029\\
Ours                  & \underline{0.988 $\pm$ 0.001} & \textbf{0.498 $\pm$ 0.089} & 0.176 $\pm$ 0.036       & 1.000 $\pm$ 0.001          & \textbf{0.403 $\pm$ 0.030} & 0.261 $\pm$ 0.027\\
\midrule
Oracle                & 0.989 $\pm$ 0.002          & 0.991 $\pm$ 0.003          & 0.001 $\pm$ 0.000          & 1.000 $\pm$ 0.001          & 0.703 $\pm$ 0.145          & 0.065 $\pm$ 0.030\\
\bottomrule
\end{tabular}%
}
\end{table}

\paragraph{Identifiability diagnostic.}
Beyond aggregate accuracy, we test whether the structured posterior
recovers the ground-truth concept map across initialisations.
\Cref{tab:identif} reports the number of near-oracle seeds (structured
concept F1 above $0.85$) and the number of distinct concept-mapping
clusters across seeds (K-means with silhouette selection, reported
as $n_{\mathrm{modes}}$). The
structured posterior of Ours reaches a near-oracle concept map in
$9/10$ runs, while DPL$^\ddagger$, SPL$^\dagger$, HardPC$^\star$,
and BEARS$^\S$ (k\!=\!1) do not reach the near-oracle threshold in
any run. The raw perception head improves less
than the structured decoder, so this diagnostic is evidence that the
posterior resolves the targeted shortcut in most runs, not that the neural
head alone fully recovers oracle concepts. The pattern is consistent
with the shortcut-weighting analysis: the level-$3$ ordering prior
penalises the position-permutation shortcut during non-degenerate
training and structured decoding.

\begin{table}[t]
\centering
\caption{Concept identifiability diagnostic on MNIST-Add.}
\label{tab:identif}
\begin{tabular}{lcccc}
\toprule
Method & F1(C) raw & F1(C) structured & near-oracle & $n_{\mathrm{modes}}$\\
\midrule
DPL$^\ddagger$       & $0.215\pm0.110$ & $0.222\pm0.108$ & $0/10$ & $3$\\
SPL$^\dagger$        & $0.213\pm0.076$ & $0.208\pm0.081$ & $0/10$ & $2$\\
HardPC$^\star$       & $0.267\pm0.070$ & $0.257\pm0.074$ & $0/10$ & $2$\\
BEARS$^\S$ (k\!=\!1) & $0.215\pm0.112$ & $0.210\pm0.122$ & $0/10$ & $2$\\
Ours                 & $\mathbf{0.498\pm0.089}$ & $\mathbf{0.915\pm0.199}$ & $\mathbf{9/10}$ & $2$\\
\bottomrule
\end{tabular}
\end{table}

\subsection{Ablation study}
\label{sec:ablation}

The ablations show that the gain is not caused by adding more circuit
structure but by a specific prior signal and a specific objective
component.

\paragraph{Prior informativeness and strength.}
On MNIST-Add (\Cref{tab:ablation}, left), the largest drop occurs
when the ordering prior is removed. Further removing parity leaves
performance close to the empty-prior setting, so the remaining
task-schema constraints do not recover the lost concept information.
The $\beta$ sweep (\Cref{fig:noise-beta}, left) gives the same
conclusion under a fixed circuit: concept recovery increases as the
ordering prior receives more weight, then saturates once the
exponential reweighting already separates the main shortcut maps. The
gain therefore comes from the prior that resolves the position-
permutation symmetry, not from adding more constraints.

\paragraph{Objective and support choices.}
On BDD-OIA-6c (\Cref{tab:ablation}, right), removing the
$Z_{\mathrm{all}}$ denominator (substituting the unnormalised
surrogate $-\log Z_y$) causes the largest concept-F1 collapse even
though task accuracy stays flat.
Disabling soft penalties ($\beta\!=\!0$) and hardening every expert
axiom into the support both produce smaller drops; the hard-support
variant performs worse than the soft version, matching the main
comparison ordering. Task accuracy stays in the same narrow band
across all variants, so the ablation isolates concept recovery rather
than label fitting.

\begin{table}[t]
\centering
\caption{\textbf{Ablation studies.}
Left: prior-richness ablation on MNIST-Add.
Right: component ablation on BDD-OIA-6c.}
\label{tab:ablation}
\resizebox{\linewidth}{!}{%
\begin{tabular}{l cc | l cc}
\toprule
\multicolumn{3}{c|}{\textit{Prior-richness (MNIST-Add)}} &
\multicolumn{3}{c}{\textit{Component (BDD-OIA-6c)}} \\
Variant & F1(Y)\,$\uparrow$ & F1(C)\,$\uparrow$ 
&
Variant & Acc.(Y)\,$\uparrow$ & F1(C)\,$\uparrow$ 
\\
\midrule
Level~3 ($=$ Ours)     & \textbf{0.988} $\pm$ 0.001 & \textbf{0.498} $\pm$ 0.089 
& Ours                     & \textbf{0.768} $\pm$ 0.002 & \textbf{0.362} $\pm$ 0.001 
\\
Level~2 ($-$ ordering) & 0.986 $\pm$ 0.003 & 0.260 $\pm$ 0.072 
& w/o soft ($\beta\!=\!0$) & 0.767 $\pm$ 0.001 & 0.332 $\pm$ 0.041 
\\
Level~1 ($-$ parity)   & 0.986 $\pm$ 0.004 & 0.211 $\pm$ 0.068 
& w/o normalisation        & 0.766 $\pm$ 0.001 & 0.113 $\pm$ 0.004 
\\
Level~0 (empty prior)  & 0.986 $\pm$ 0.005 & 0.222 $\pm$ 0.088 
& hard-support variant     & 0.767 $\pm$ 0.001 & 0.336 $\pm$ 0.027 
\\
\bottomrule
\end{tabular}%
}
\end{table}

\subsection{Robustness to noisy ontologies}
\label{sec:exp-noise}

\begin{figure}[!t]
\centering
\begin{minipage}[c]{0.32\linewidth}
  \includegraphics[width=\linewidth]{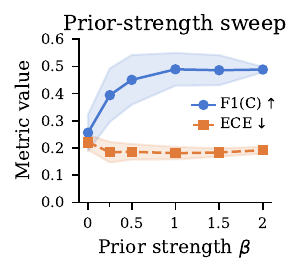}
\end{minipage}\hfill
\begin{minipage}[c]{0.32\linewidth}
  \includegraphics[width=\linewidth]{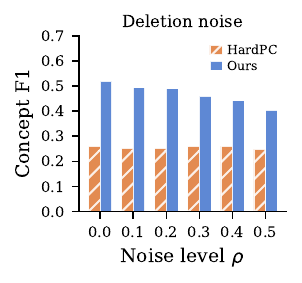}
\end{minipage}\hfill
\begin{minipage}[c]{0.32\linewidth}
  \includegraphics[width=\linewidth]{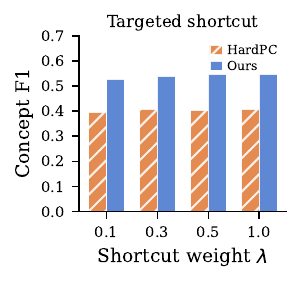}
\end{minipage}
\caption{\textbf{Prior strength and ontology noise on MNIST-Add.}
Left: prior-strength sweep, reporting raw-$\theta$ concept F1 and ECE
as $\beta$ varies.
Middle: structured concept F1 under random deletion of soft prior
axioms at corruption fraction $\rho$.
Right: structured concept F1 under targeted shortcut injection as
the injected axiom weight $\lambda$ increases; $\lambda\!=\!0$
reuses the clean-prior baseline of \Cref{tab:identif}.}
\label{fig:noise-beta}
\end{figure}

We probe two contrasting forms of soft-prior corruption on MNIST-Add
and evaluate their effect on the structured concept decoder. Random
deletion of soft prior axioms (\Cref{fig:noise-beta}, middle) causes
a gradual decline in structured concept F1 as the corruption fraction
$\rho$ increases. This behaviour is consistent with
Proposition~\ref{prop:noise}: bounded changes to soft axiom weights
perturb the posterior continuously. Targeted shortcut injection
(\Cref{fig:noise-beta}, right) is the adversarial counterpart. As the
weight of the injected shortcut axiom increases, the structured
decoder is biased toward the shortcut configuration, even though
violating worlds remain in the support. This targeted corruption is
outside the bounded random-deletion model of
Proposition~\ref{prop:noise}; it is the main failure mode of soft
compilation and motivates learning axiom reliabilities
or gating axioms between soft and hard use.

\FloatBarrier
\section{Conclusion}
\label{sec:conclusion}

We introduced soft ontology priors for weakly supervised NeSy. The
label-normalised WMC objective compiles hard support and soft penalties
into tractable posterior inference. Empirically, targeted ontology
information improves concept recovery on BDD-OIA-6c and resolves most
MNIST-Add shortcut modes, while task accuracy alone remains insensitive.

The main limitation is ontology reliability. Random deletion noise is
tolerated, but targeted shortcut axioms can still bias the decoder;
fixed weights and one global $\beta$ cannot identify unreliable or
incomplete knowledge, which can preserve the wrong concept
explanation. Since this remains label-only, concept labels or
shortcut-aware perception models may still provide stronger signal,
especially OOD, where task labels can underconstrain concepts and leave
shortcuts unresolved. Future work should learn axiom reliability and
combine ontology priors with those supervision signals.

\bibliography{nesy2026-sample}

\newpage
\appendix

\section{Proofs}
\label{apd:proofs}

\subsection{Proof of Proposition~\ref{thm:leaf-fold}}
For $Z_y$, the jointly compiled or structurally compatible circuit
$\Cir^{\Out}_{\Onto,\Prog}$ enumerates exactly the worlds satisfying
both the hard ontology support and $\Prog(\Concepts)=\Out$, while its
fixed weights encode the soft score
$\exp(-\beta\sum_j\lambda_jv_j(\Concepts))$.  After injecting neural
literal evidence, its root value is
\[
   \sum_{\Concepts\models\Cir^{\Out}_{\Onto,\Prog}}
   p_{\Parm}(\Concepts\mid\Inp)\,
   \widetilde p_{\Onto}(\Concepts;\beta)=Z_y(\Inp).
\]
The same argument on $\Cir_{\Onto}$ without the task restriction gives
$Z_{\mathrm{all}}(\Inp)$.  Independence of $p_{\Parm}$ allows the
neural term to be attached locally to literal occurrences;
decomposability propagates these factors through products over
disjoint scopes; smoothness aligns branch scopes at sums; and
determinism avoids double-counting worlds.  Each circuit is evaluated
by one bottom-up pass, giving the stated
$\mathcal{O}(|\Cir^{\Out}_{\Onto,\Prog}|+|\Cir_{\Onto}|)$ runtime.
Differentiability follows from the log-space product and
$\mathrm{logsumexp}$ recurrences.\hfill$\square$

\subsection{Proof of Proposition~\ref{prop:count}}
For a fixed label-preserving deterministic map $\alpha$, the
label-restricted ontology score on one example is
\[
  \mathbf{1}[\Prog(\alpha(\Concepts))=\Out]\,
  \exp\!\left(-\beta\sum_j\lambda_jv_j(\alpha(\Concepts))\right).
\]
The indicator is $1$ on the data support by definition of
$\mathcal{A}$. Taking population expectations under $p^\star$ gives
\[
  \log S_\Onto(\alpha)
  =-\beta\,\mathbb{E}_{p^\star}\!\left[
     \sum_j\lambda_jv_j(\alpha(\Concepts))\right]
  =-\beta V_\Onto(\alpha),
\]
up to constants common to all label-preserving maps. Taking the
ratio of two such scores gives the stated exponential
down-weighting. This proof deliberately concerns the label-restricted
mass (or, equivalently, the structured posterior before taking the
exact one-hot limit). If the normalised objective
$\log Z_y-\log Z_{\mathrm{all}}$ is evaluated after $p_{\Parm}$ has
become an exact point mass at a label-preserving assignment, then the
same assignment is the only non-zero term in both partitions and the
two soft scores cancel. Thus the proposition should not be read as
claiming non-zero conditional-likelihood odds at the degenerate
vertex; it describes the prior-induced score and gradient bias that
acts during non-degenerate training and structured decoding. In the
hard limit $\beta\!\to\!\infty$, any positive-violation map receives
zero label-restricted support.\hfill$\square$

\subsection{Proof of Proposition~\ref{prop:noise}}
Write $p_{\Onto}(\Concepts;\beta)=
Z(\Onto;\beta)^{-1}\exp(-\beta E_{\Onto}(\Concepts))$ with partition
$Z(\Onto;\beta)=\sum_{\Concepts}\exp(-\beta E_{\Onto}(\Concepts))$,
and analogously for $\Onto'$. The assumption
$|E_{\Onto}(\Concepts)-E_{\Onto'}(\Concepts)|\le\varepsilon$ implies
$|{-}\beta E_{\Onto}+\beta E_{\Onto'}|\le\beta\varepsilon$ pointwise,
hence $|\log Z(\Onto;\beta)-\log Z(\Onto';\beta)|\le\beta\varepsilon$
by the standard log-sum-exp bound, and therefore
$|\log p_{\Onto}(\Concepts;\beta)-\log p_{\Onto'}(\Concepts;\beta)|
\le 2\beta\varepsilon$, which is the log-density perturbation bound of
Proposition~\ref{prop:noise}. The KL bound follows by
$\mathrm{KL}(p_{\Onto}\|p_{\Onto'})
=\mathbb{E}_{p_{\Onto}}[\log p_{\Onto}-\log p_{\Onto'}]
\le 2\beta\varepsilon$. The per-axiom noise model satisfies
$\varepsilon=\mathcal{O}(\rho K\Delta)$ because at most $\rho K$
axioms change and each changes the per-world cost by at most $\Delta$.
For the hard limit, the support is the indicator
$\mathbf{1}[E_{\Onto}(\Concepts)=0]$, which is not continuous in
$\Onto$: a single flipped axiom $a_j$ that excludes a world $\Concepts$
with $p^\star(\Concepts)>0$ sends $p_{\Onto'}(\Concepts)=0$ while
$p_{\Onto}(\Concepts)>0$, so the KL diverges and the gradient on
$p_{\Parm}$ at that world vanishes.\hfill$\square$

\section{Compilation Details}
\label{apd:compile}

\begin{algorithm}[h]
\floatconts{alg:forward}
{\caption{Forward / backward pass for one training example.}}
{%
\DontPrintSemicolon
\KwIn{Ontology circuit $\Cir_{\Onto}$; label-restricted joint circuit
       $\Cir^{\Out}_{\Onto,\Prog}$; perception model $p_{\Parm}$;
       input $\Inp$; label $\Out$.}
\KwOut{$-\log Z_y(\Inp)+\log Z_{\mathrm{all}}(\Inp)$ with
       $\nabla_{\Parm}$ populated.}
$\boldsymbol{\theta} \leftarrow p_{\Parm}(\cdot\mid\Inp)$\;
\ForEach{literal occurrence $(c_i\!=\!v)$ in both circuits}{
  $e_i^v \leftarrow v\log\theta_i + (1-v)\log(1-\theta_i)$
}
$\log Z_y \leftarrow \mathrm{LogWMC}\bigl(\Cir^{\Out}_{\Onto,\Prog};\{e_i^v\}\bigr)$\;
$\log Z_{\mathrm{all}} \leftarrow \mathrm{LogWMC}\bigl(\Cir_{\Onto};\{e_i^v\}\bigr)$\;
\tcp*{$+$ at \textsc{Product}, $\mathrm{logsumexp}$ at \textsc{Sum}; fixed ontology weights stay in the circuit}
\Return $-\log Z_y+\log Z_{\mathrm{all}}$\;
}%
\end{algorithm}

\paragraph{Per-axiom compilation.}
Each axiom is compiled by enumerating its local scope; the
resulting sd-DNNF fragment has $|S_a|+1$ levels: one level of leaves
per variable, one level of products per satisfying assignment, and a
single top-level sum.  The fragment size is $\mathcal{O}(2^{|S_a|}\,|S_a|)$.

\paragraph{Connected-component conjunction.}
Axioms sharing variables form a connected component in the
variable-sharing graph.  For each component we enumerate the shared
scope and emit a single sum-of-products that satisfies every axiom in
the component.  Components with disjoint scopes are combined under a
top-level product, which preserves decomposability.  Variables that
no axiom mentions are appended as smoothing literals to keep the
circuit smooth over $[n]$.

\paragraph{Task--ontology conjunction.}
Re-enumerate the task circuit's product terms and filter them through
the ontology circuit, fusing the per-leaf log-prior offsets in the
process.  The resulting sd-DNNF has size at most
$|\Cir_{\Onto}|\cdot|\Cir_{\Prog}^{\Out}|$ but is typically far smaller
because most task terms are ontology-consistent.

\paragraph{Materialised circuit sizes.}
The circuits used in the experiments are small after task-specific
enumeration. In an unshared sum-of-products upper bound, BDD-OIA-6c
uses at most $64$ all-label worlds over six concepts; MNLogic uses
$8$ worlds over three concepts; MNIST-Half uses $25$ digit pairs with
at most $5$ pairs for a single sum; and MNIST-Add uses $100$ digit
pairs with at most $10$ pairs for a single sum. Counting one literal
per concept variable per product, the largest all-label MNIST-Add
circuit has about $100\times20$ literal occurrences plus $100$
product nodes and one sum node; the label-restricted circuit has at
most about $10\times20$ literal occurrences. Thus the WMC part is a
few thousand log-space operations per example in these benchmarks and
was not the memory bottleneck relative to the neural forward/backward
passes on the hardware reported in \Cref{sec:exp-setup}.

\section{Datasets and Prior Specifications}
\label{apd:data}

\paragraph{BDD-OIA-6c concept vocabulary and stop rule.}
The canonical traffic view uses six binary concepts: $c_0$--$c_5$
respectively denote red light on, green light on, pedestrian present,
rider present, vehicle ahead, and clear path. They are derived from the
21-column RSBench preprocessed labels by the audited disjunction
mapping in the code release. The
binary stop label $y\in\{0,1\}$ is the deterministic stop rule
$\Prog(\Concepts)=\neg[\,(c_1\vee c_5)\wedge\neg c_0\wedge\neg c_2
\wedge\neg c_3\wedge\neg c_4\,]$:
go (i.e.\ $y\!=\!0$) requires at least one forward cue
($c_1=$~green\_light or $c_5=$~clear\_path) active and all four
stop-blocking concepts ($c_0,c_2,c_3,c_4$) inactive. The mapping is
gated by both an ontology-alignment audit (each candidate expert
axiom must have a violation rate below $5\%$ on the real concept
labels) and a stop-rule label audit (rule-derived stop labels must
agree with the RSBench action label on $\geq 95\%$ of the test set);
we report a pre-training rule--label agreement of $0.971$.

\paragraph{MNIST-Add ontology richness levels.}
We compile four richness levels over the $2\!\times\!10$ digit
one-hot vocabulary. \emph{L0}: empty ontology, corresponding to the
no-prior task-circuit setting. \emph{L1}: one-hot constraints
\textsc{ExactlyOne}$(d_{p,0},\dots,d_{p,9})$ for each digit position
$p\in\{1,2\}$. \emph{L2}: L1 plus two parity auxiliary concepts per
position with hard \textsc{SubClass} axioms
$d_{p,d}\!\sqsubseteq\!\textit{even}_p$ for even $d$ and
$d_{p,d}\!\sqsubseteq\!\textit{odd}_p$ for odd $d$, plus the
\textsc{DisjointPair}$(\textit{even}_p,\textit{odd}_p)$ for each
position. \emph{L3}: L2 plus a soft ordering prior implemented as
non-negative \textsc{DisjointPair}$(d_{1,i},d_{2,j})$ weights for
every pair with $i>j$; the per-pair weight is
$1/\text{ordering\_scale}$ before $\beta$ scaling. Pairs are
constructed with the \emph{sorted-pair} option (smaller digit assigned
to position~1) so that the L3 prior aligns with the data-generating
process.

\paragraph{MNIST-Half splits and ontology.}
The five-digit alphabet $\{0,\dots,4\}$ defines a $10$-concept
vocabulary that is identical in shape to MNIST-Add but smaller; the
ontology richness levels L0--L3 are the MNIST-Add definitions above
restricted to five digits per position. The in-distribution unordered
pair set $\{(0,0),(0,1),(2,3),(2,4)\}$ expands to the seven ordered
pairs $\{(0,0),(0,1),(1,0),(2,3),(3,2),(2,4),(4,2)\}$ that train,
validation, and in-distribution test draw from; the OOD split draws
from the eighteen complementary ordered pairs of $\{0,\dots,4\}^2$.

\paragraph{MNLogic prior and splits.}
The benchmark-provided in-distribution combination set
$\{001,010,000\}$ over $(a,b,d)$ is encoded as the DIMACS CNF
$(\neg a)\wedge(\neg b\vee\neg d)$, equivalent to
$\neg a\wedge\neg(b\wedge d)$. The prior is treated as a soft
distribution prior (\Cref{sec:method-prior},
$\SoftAxioms_{\mathrm{bench}}$): each train/validation/test sample
is drawn from an in-distribution combination with probability $0.8$
and from any other combination with probability $0.2$, so $20\%$ of
the training data violates the CNF and a hard support encoding would
be infeasible. The OOD split conditions the draw on the seven
remaining combinations.

\paragraph{Split sizes.}
\Cref{tab:split-sizes} lists the per-split sample counts for every
benchmark. BDD-OIA-6c uses the RSBench preprocessed train/val/test
release verbatim; MNIST-Add and MNIST-Half generate fresh pairs per
seed within the listed bounds; MNLogic generates fresh triples per
seed at the listed sizes.

\begin{table}[h]
\centering
\caption{Per-split sample counts.}
\label{tab:split-sizes}
\begin{tabular}{lcccc}
\toprule
Dataset & Train & Val. & Test (in-dist.) & OOD test\\
\midrule
BDD-OIA-6c       & \multicolumn{4}{c}{RSBench preprocessed release (no OOD)}\\
MNIST-Add        & $30{,}000$ & $2{,}000$ & $5{,}000$  & $2{,}000$ (sums $\geq\!9$)\\
MNIST-Half       & $30{,}000$ & $1{,}000$ & $2{,}000$  & $2{,}000$ (held-out pairs)\\
MNLogic-3bit-CNF & $30{,}000$ & $5{,}000$ & $5{,}000$  & $5{,}000$ (held-out combinations)\\
\bottomrule
\end{tabular}
\end{table}

\paragraph{Prior provenance.}
The code release stores each prior with an explicit provenance label.
MNIST-Add and MNIST-Half use task-induced digit-position schemas
(one-hot per position, with optional parity/order enrichments for
the L2/L3 levels above); these schemas are induced by the constructed
task and are not MNIST metadata. RSBench priors are split by source:
official CNF priors, when present and supported by the compiler, are
reported as benchmark-provided. BDD-OIA has two views; the primary
21-concept view is the raw-label baseline (no expert prior because
the audit on real test labels found no supported annotation
invariant), and the canonical six-concept traffic view of
\Cref{sec:exp-main} is gated by the audits above.

\section{Implementation Details}
\label{apd:impl}
The MNIST backbone is a two-conv LeNet with feature dimension $128$,
shared across the two digit positions for MNIST-Add and across the
three binary digit positions for MNLogic; BDD-OIA inputs are the
2048-d RSBench-preprocessed image features released by
\citet{bortolotti2024rsbench} and our network applies a two-layer MLP
head on top of those embeddings.
Optimisation uses Adam with learning rate $10^{-3}$ and weight decay
$0$. BDD-OIA-6c, MNIST-Add, and MNLogic train for $30$ epochs;
MNIST-Half trains for $60$ epochs following the BEARS sparse-training
protocol. Batch size is $256$ for BDD-OIA-6c and $128$ for MNIST-Add,
MNIST-Half, and MNLogic. Each epoch processes the fixed BDD-OIA
training split, $30{,}000$ randomly paired MNIST-Add samples,
$30{,}000$ randomly paired MNIST-Half samples, or $30{,}000$
generated MNLogic triples, depending on the benchmark.
All hyperparameters are specified in version-controlled configuration
files; reported sweeps specify the selected $\beta$ or prior-weight
setting in the corresponding table.

\section{Additional Experimental Results}
\label{apd:extra}

\paragraph{Out-of-distribution table.}
\Cref{tab:ood-app} reports additional OOD diagnostics on MNIST-Add.

\begin{table}[h]
\centering
\caption{Out-of-distribution diagnostics on MNIST-Add: training
restricted to sums $\leq 8$, evaluated on the held-out sums
$\{9,\dots,18\}$.}
\label{tab:ood-app}
\resizebox{\linewidth}{!}{%
\begin{tabular}{lcccc}
\toprule
Method & In-dist.\ Acc.\,$\uparrow$ & OOD Acc.\,$\uparrow$ & OOD F1(C)\,$\uparrow$ & OOD ECE\,$\downarrow$\\
\midrule
DPL$^\ddagger$ & 0.627 $\pm$ 0.022 & 0.000 $\pm$ 0.000 & 0.026 $\pm$ 0.008 & $\mathbf{0.175 \pm 0.018}$\\
SPL$^\dagger$ & 0.974 $\pm$ 0.023 & $\mathbf{0.029 \pm 0.028}$ & 0.037 $\pm$ 0.026 & 0.216 $\pm$ 0.003\\
HardPC$^\star$ & 0.990 $\pm$ 0.001 & 0.021 $\pm$ 0.018 & 0.030 $\pm$ 0.010 & 0.282 $\pm$ 0.011\\
BEARS$^\S$ & $\mathbf{0.993 \pm 0.001}$ & 0.000 $\pm$ 0.000 & 0.025 $\pm$ 0.006 & 0.180 $\pm$ 0.017\\
Ours & 0.991 $\pm$ 0.001 & 0.014 $\pm$ 0.024 & $\mathbf{0.077 \pm 0.042}$ & 0.263 $\pm$ 0.067\\
\midrule
Oracle & 0.994 $\pm$ 0.001 & 0.259 $\pm$ 0.068 & 0.281 $\pm$ 0.084 & 0.063 $\pm$ 0.005\\
\bottomrule
\end{tabular}%
}
\end{table}

\begin{table}[h]
\centering
\caption{BDD-OIA-6c prior-strength sweep for Ours under the canonical
traffic prior; best value per row in \textbf{bold}.}
\label{tab:bdd-beta-app}
\resizebox{\linewidth}{!}{%
\begin{tabular}{lcccccc}
\toprule
\(\beta\)        & $0$          & $0.25$       & $0.5$        & $1$          & $2$          & $4$\\
\midrule
F1(C)$\uparrow$  & $0.332\pm0.041$ & $0.361\pm0.000$ & $0.361\pm0.001$ & $\mathbf{0.362\pm0.001}$ & $0.361\pm0.001$ & $0.362\pm0.001$\\
Acc.(Y)$\uparrow$ & $0.767\pm0.001$ & $0.766\pm0.001$ & $0.765\pm0.002$ & $\mathbf{0.768\pm0.002}$ & $0.765\pm0.001$ & $0.766\pm0.001$\\
ECE$\downarrow$  & $\mathbf{0.343\pm0.007}$ & $0.346\pm0.007$ & $0.347\pm0.005$ & $0.351\pm0.001$ & $0.351\pm0.001$ & $0.351\pm0.003$\\
\bottomrule
\end{tabular}
}
\end{table}

\paragraph{Concept-supervision sweep.}
\Cref{tab:absweepc-app} reports the effect of adding sparse concept
supervision, complementing the label-only results in
\Cref{sec:ablation}. Even $\lambda_{\mathrm{sup}}\!=\!1$ already far
exceeds the best ontology-prior-only result (Ours-L3, raw F1 $0.498$).

\begin{table}[h]
\centering
\caption{Concept-supervision sweep on MNIST-Add at
$L\!=\!1,\beta\!=\!1$.}
\label{tab:absweepc-app}
\begin{tabular}{ccc}
\toprule
$\lambda_{\mathrm{sup}}$ & F1(C)\,$\uparrow$    & Acc.\,$\uparrow$\\
\midrule
$0.0$                    & $0.211 \pm 0.068$    & $0.986$\\
$0.1$                    & $0.282 \pm 0.067$    & $0.976$\\
$1.0$                    & $\mathbf{0.895 \pm 0.000}$ & $\mathbf{0.994}$\\
$10.0$                   & $\mathbf{0.895 \pm 0.001}$ & $\mathbf{0.994}$\\
\bottomrule
\end{tabular}
\end{table}

\paragraph{Raw-$\theta$ vs.\ structured-posterior concept F1.}
\Cref{tab:raw-vs-struct} reports both concept-F1 variants of
\Cref{sec:exp}. The raw-$\theta$ column (our primary metric) scores the
perception head directly; the structured column scores the circuit
argmax given the model's own predicted task label. The two nearly
coincide on BDD-OIA-6c (except for the BEARS ensemble, whose
inference-time argmax differs), where a single task bit barely
constrains the six concepts. On MNIST-Add the predicted $19$-valued sum strongly
constrains the digit pair, so structured decoding is far higher for
every circuit-equipped method, most strikingly Ours, whose
informative ordering prior lets the circuit recover the digits almost
perfectly at decode time even though the head alone remains partly
shortcut. We read this gap as the decoding capability of the symbolic
layer, and report the harder raw-$\theta$ metric in the main tables to
avoid crediting concept recovery to label-conditioned decoding.

\begin{table}[h]
\centering
\caption{Raw versus structured concept F1. The structured score uses
the circuit MAP given the predicted label.}
\label{tab:raw-vs-struct}
\begin{tabular}{l cc cc}
\toprule
& \multicolumn{2}{c}{BDD-OIA-6c} & \multicolumn{2}{c}{MNIST-Add}\\
\cmidrule(lr){2-3}\cmidrule(lr){4-5}
Method & raw-$\theta$ & structured & raw-$\theta$ & structured\\
\midrule
DPL$^\ddagger$       & $0.309$ & $0.309$ & $0.215$ & $0.222$\\
SPL$^\dagger$        & $0.279$ & $0.279$ & $0.213$ & $0.208$\\
HardPC$^\star$       & $0.330$ & $0.330$ & $0.267$ & $0.257$\\
BEARS$^\S$ (k\!=\!3) & $0.291$ & $0.184$ & $0.240$ & $0.314$\\
Ours                 & $0.361$ & $0.361$ & $0.498$ & $0.915$\\
\midrule
Oracle               & $0.456$ & $0.454$ & $0.991$ & $0.933$\\
\bottomrule
\end{tabular}%
\end{table}

\paragraph{BDD-OIA.}
The code release evaluates a binary stop/go task circuit over the
RSBench-preprocessed BDD-OIA features~\citep{bortolotti2024rsbench}.
The main BDD-OIA setting uses the audited canonical six-concept traffic
view and the expert-authored traffic prior.  The mapping audit also
checks the canonical stop rule against the released BDD-OIA stop
labels; the formal runs report rule-label accuracy $0.971$ before
training.  The primary 21-concept view is retained only as a no-prior
raw-label baseline: the audit on real test labels rejects unsupported
annotation invariants on this release, so we keep this view empty
rather than encode unaudited expert claims.

\end{document}